# ADABOOK & MULTIBOOK
## *Adaptive Boosting with Chance Correction*

David M W Powers[1,2]

[1] *Beijing Municipal Lab for Multimedia & Intelligent Software, Beijing University of Technology, Beijing, China*
[2] *CSEM Centre for Knowledge & Interaction Technology, Flinders University, Adelaide, South Australia*
powers@ieee.org

Keywords: Kappa, Correlation, Informedness, Markedness, ROC, AUC, Boosting, Bagging, AdaBoost, MultiBoost, Machine Learning, Signal Processing, Classifier Fusion, Language Technology, Concept Learning.

Abstract: There has been considerable interest in boosting and bagging, including the combination of the adaptive techniques of AdaBoost with the random selection with replacement techniques of Bagging. At the same time there has been a revisiting of the way we evaluate, with chance-corrected measures like Kappa, Informedness, Correlation or ROC AUC being advocated. This leads to the question of whether learning algorithms can do better by optimizing an appropriate chance corrected measure. Indeed, it is possible for a weak learner to optimize Accuracy to the detriment of the more reaslistic chance-corrected measures, and when this happens the booster can give up too early. This phenomenon is known to occur with conventional Accuracy-based AdaBoost, and the MultiBoost algorithm has been developed to overcome such problems using restart techniques based on bagging. This paper thus complements the theoretical work showing the necessity of using chance-corrected measures for evaluation, with empirical work showing how use of a chance-corrected measure can improve boosting. We show that the early surrender problem occurs in MultiBoost too, in multiclass situations, so that chance-corrected AdaBook and Multibook can beat standard Multiboost or AdaBoost, and we further identify which chance-corrected measures to use when.

## 1 INTRODUCTION

Boosting is a technique for turning a weak learner into a strong learner in terms of Valiant's (1984,1989) Probably Approximately Correct framework, where a strong learner is defined informally as being arbitrarily close to perfect and a weak learner is defined informally as being marginally better than chance, where the performance of the algorithms is limited as a polynomial of the reciprocals of the arbitarily small deviations from perfection or chance respectively. However, Shapire's (1989) original algorithm and proof for boosting only considered the dichotomous (two class) case and made the assumption that chance level performance was $1/2$ on the basis that guesses are unbiased coin tosses. Practical boosting algorithms (Freund, 1995) followed based on the same idea an iteratively applied weak learner, concentrating on the examples which were not classified correctly.

Adaptive boosting (Freund & Shapire, 1997), AdaBoost, used weights on instances for the next training of the weak learner were adjusted according to the odds $a/e$ of being correct in order to even up the score and force finding a new way of making an above chance decision, where $a$ is the accuracy (proportion right) and $e$ is the error (proportion wrong). In returning the composite classifier, a linear weighting using the log odds is used: $\ln(a/e)$. While $1/2 < a < 1$ boosting can continue – otherwise the final classifier is built: equality with 1 means that the weak learner returned a perfect result and the stronger learner has been successfully achieved, while equality with $1/2$ means that a chance level score was achieved and the weak learner has failed.

Many generalizations exist to the multiclass case (Shapire & Freund, 2012), including a variant on the dichotomous algorithm that simply used a K-class learner but retained the far too strong $1/2$ weak learning threshold (M1), and a variant that estimates pseudo-loss instead of error, based on modified weak learners that return plausibility estimates of the classes (M2). However SAMME attempts to replace the $e < 1/2$ or $a > 1/2$ condition by an $a > 1/K$ condition (Zhu et.al.,2009), but this is still problematic as after reweighting the true chance level may be considerably different from this. If a node has just 2 of K classes, SAMME still accepts any accuracy over $1/K$.

However, there is another problem that affects K=2 as well, and relates to a growing concern with simple measures of accuracy and error, and has led to the proposal and use of chance-corrected measures, including in particular the various forms of Kappa and the probability of an informed decision, Informedness (Powers, 2003, 2011): it is possible to make your learner dumber and worser but get higher accuracy! In all forms of boosting, the effective distributions of examples varies as it concentrates on the poorly performing instances. But if we have a 60:40 prevalence of one class, then $a > 1/K$ is trivial to achieve for any K≥2 simply by guessing the majority class. In fact, the booster may just rebalance the prevalences to offset this bias, but in doing so perform many needless iterations.

This therefore raises the questions of whether we can ensure that a learner is optimized for chance-corrected performance rather than accuracy, which existing learners have this property, and whether boosting will perform better if it boosts based on a chance corrected measure rather than accuracy. We will review chance-corrected measures in the next section and assume a basic familiarity with the standard Rand Accuracy and Information Retrieval measures, but Powers (2011) provides a thorough review of both corrected and uncorrected measures.

We note that in this paper we use the new statistics rather than the deprecated Statistical Hypothesis Inference Testing (Cummings, 2012), viz. showing effects graphically rather than tabulating with p-values or alphas, providing standard deviations to allow understanding of the variance and bias of the approaches, and confidence intervals of two standard errors to allow understanding of the reliability of the estimated effects sizes. We also note that we avoid displaying or averaging accuracies (or F-scores), which are incomparable unless biases and prevalences are matched (Powers, 2011,2012). However these results themselves can look much better and it favours our proposed algorithms even more if we use these traditional but unsound and deprecated measures!

Boosting Accuracy (as performed by standard Adaboost) need *not* boost a chance corrected measure and may *not* even satisfy weak learnability even though Accuracy appears to – it merely satisfies 2-learnability which is the surrogate used in the proof of Adaboost (Freund & Shapire,1997). On the other hand, boosting the appropriate chance-corrected measure can in general be seen to improve Accuracy and F-score, and we will take the opportunity to note places where the base learner does very poorly, and Accuracy or F1 rises, but the chance-corrected measures actually fall.

## 1.1 Applications: Text & Signal Processing

Whereas the preceding discussion has been in a general Machine Learning context we take a moment to bring the discussion to the practical level and discuss the kinds of applications and learning algorithms where the proposed techniques can make a huge difference. The practical context of the work reported here, including the development and testing of the various chance-corrected measures, is signal processing of Electroencephalographic Brain Computer Interface experiments (Fitzgibbon et al. 2007,2013; Atyabi et al., 2013), Audiovisual Speech, Gesture, Expression and Emotion Recognition (Lewis & Powers, 2004; Jia et al. 2012,2013) and Information Retrieval and Language Modelling (Powers, 1983, 1991; Yang & Powers, 2006; Huang & Powers, 2001), and it was in this Natural Language Processing context that the problems of evaluation and its roles in the misleading of learning systems were first recognized (Entwisle & Powers, 1998). Furthermore, although the work reported here uses standard datasets, we use character/letter datasets that pertain to this Natural Language task because the problem we are identifying, and the advantage of solving it, grows with the number of classes. Similarly the software used is modifications of standard algorithms as implemented in Weka (Witten et al., 2011), so that comparison with other multiclass boosting work is possible. Two of Weka's Boosting algorithms, AdaBoostM1and MultiBoost (Webb,2000), are used as the basis for the proposed modifications, with Tree-based and Perceptron-based learners preferred to stable learners, like Naïve Bayes, that don't boost (Fig. 1).

There are two practical considerations that make these two boosting algorithms (and the whole family of boosting algorithms based on equivalently error <½ or accuracy >½) unsuited for signal processing and classification in our real world applications: the multiclass nature of the work (unsuccessfully addressed by in the cited work, and the high dimensional noisy data (unrelated to the traditional label noise model).

## 2 CHANCE CORRECTION

Several variants of chance-correction exist, with a family of accuracy correction techniques, Kappa κ in terms of the Accuracy *a* (which is usually Rand Accuracy, but can be Recall or Precision or any other probability-like measure, and we illustrate these in terms of counts of various conditions or contingencies):

$$\kappa = [a-\hat{E}(a)]/[1-\check{E}(a)]$$
$$a_{Rand} = \text{AllCorrect/AllCases}$$
$$a_{Recall} = \text{TruePositives/RealPos}$$
$$a_{Precision} = \text{TruePositives/PredictPos}$$

(1)

The expected accuracy is defined differently depending on the particular method of chance correction, the common version dropping the hats and using the same value of E(*a*) top and bottom, but Cohen's Kappa using a different definition of expectedness from the Scott and Fliess versions, with "Bookmaker" Informedness being shown to satisfy a similar definition (Powers, 2011) having *different* chance estimates in numerator and denominator, its fundamental property being its definition as the probability of making an *informed* decision, with other properties relating to the dichotomous measues ROC AUC, Gini, DeltaP', as well as the empirical strength of association in psychology (Powers, 2011), with Dichotomous Informedness or DeltaP' given by:

$\hat{E}(a_{Inform}) = E(a_{Recall}) = \text{Bias} = \text{PredictedPos/AllCases}$
$\check{E}(a_{Inform}) = E(a_{Precision}) = \text{Prev} = \text{RealPos/AllCases}$   (2)
Informed = $\text{Recall}^+ + \text{Recall}^- - 1 = [\text{Recall-Bias}]/[1-\text{Prev}]$

These equations illustrate clearly why chance-corrected measures are needed, as Recall follows Bias (guess 100% positive and get 100% Recall) while Precision follows Prevalence (positives are common at 90% while negatives are rare at 10% means Precision is an expected 90% by guessing). Informedess is basically the same formula as the one used to eliminate the effect of chance from multiple choice examples. While the Kappa definition captures directly the idea of correction of an accuracy measure by subtracting off its expected value, and renormalizing to the form of a probability, only the Informedness form has a clear probablistic interpration, although they are all loosely referred to as probabilities. In the Kappa form equation for Informedness, we can understand the demoninator in terms of the room for improvement above the chance effect attributable to Prevalence (in our example we can only improve from 90% to 100%). In the dichotmous (2 class) case, measures equivalent to Informedness have been developed under various guises by a variety of different researchers as reviewed by Powers (2011), but here we merely note that the version that sums Recall for +ves and –ves can be easily related to ROC (being the height of the system above the chance line as tpr=Recall$^+$ and fpr=1–Recall$^-$).

In this paper we focus on the multiclass case (K>2 classes) using the most commonly used (Cohen) Kappa and (Powers) Informedness measures. However, all Kappa and Informedness variants, including Powers' (2011) Markedness and Matthews' Correlation, give a probability or score of 0 for chance-level performance, and 1 for perfect performance, taking values on a [-1,+1] scale as they can be applied to problems where higher or lower than chance performance is exhibited. But for ease of substitution in the various boosting algorithms it is convenient to remap the zero of these double-edged "probabilities" to a chance level of $^1/_2$ on a [0,1] scale. In fact, in the dichotomous case, Gini, and the single operating point ROC, correspond to such a renormalization of Informedness. As all of the chance=0 measures we consider can be related to the Kappa definition, we refer to these generically as Kappa, while for any accuracy like measure on a [0,1] scale we refer to it as Accuracy.

We thus define a corresponding Accuracy

$$a_\kappa = (\kappa+1)/2 \qquad (3)$$

for any chance-corrected Kappa $\kappa$, and we define the associated Error as 1 – Accuracy

$$e_\kappa = (1-\kappa)/2 \qquad (4)$$

AdaBoost and many other boosting algorithms are defined in terms of Rand Accuracy or equivalently the proportion of Error, and can thus be straight-forwardly adapted by substituting the corresponding alternate definition. Our prediction is that optimizing Cohen's Kappa and Powers' Bookmaker Informedness are expected to do far better than the uncorrected Rand Accuracy (or proportional Error) or other tested measures including Powers' Markedness and Matthews' Correlation, when tested in Weka's implementation (Witten et al., 2011) of AdaBoost.M1 using tree stumps/learners as the weak learners.

Informedness is expected to perform best if the weak learner is unbiased or prevalence-biased, but sometimes Kappa can be expected to be better, in particular, when the weak learner optimizes Kappa or Accuracy, which is linearly related to Kappa for a fixed estimate of the expected accuracy (which for Cohen Kappa corresponds to fixed marginal probabilities). Kappa can go up and Informedness down, when the predictive bias (proportion of predictions) for a particular label varies from population prevalence (proportion of real labels) (Powers, 2012). No learners that explicitly optimize Informedness are known, but all learners that match Label Bias to Class Prevalence will maximize all forms of Kappa and Correlation, including Informedness, whatever form of error they minimize or accuracy they maximize subject to that constraint. This has long been a heuristic for the setting of thresholds in neural networks, and can also be used in Receiver Operating Characteristics (ROC) optimization. However in ROC Analysis this corresponds with intersecting the fn=fp diagonal and is not in general the optimum operating point. Mismatching Bias and Prevalence can lead to gains over equal Bias and Prevalence (Powers, 2011).

Moreover, the base learners originally used with AdaBoost were tree-type learners, and in a leaf node these algorithms can be expected to make the majority decision or an equiprobable guess. The former seems to be ubiquitous but biased by the locally conditioned prevalences of that node rather than the global population prevalences that are appropriate for optimizing a chance-corrected measure, and this in particular is inappropriate for Informedness. On the other hand, network-based learners, and Bayesian learners, do not have such a simple majority voting bias. Moreover, AdaBoost as a convex learner has strong similarities to neural networks and SVM, but a Naïve Bayes learner could provide a quite distinct behaviour and provide a weak learner that also satisfies the requirement of being a fast learner.

Note that if a weak learner doesn't have the local majority bias of a tree learner, it may not be improved by the use of Bookmaker weighting (AdaBook) or Kappa weighting (AdaKap) rather than a conventional uncorrected accuracy or error optimizing learner. This raises an empirical question about whether boosting will work with different algorithms, and whether the form of chance-correction that corresponds to our analysis and hypotheses indeed performs best.

Our second prediction is that failure of AdaBook and AdaKap and standard AdaBoost with Rand Accuracy, can be expected at times, with "early stopping" due to the weak learner failing to satsify the $1/2$ condition. But for different distributions, and different weak learner optimization criteria, one can improve and another worsen.

Multiboost (Webb,2000) seeks to avoid this "early stopping" by interleaving bagging amongst the Adaboost iterations – we use the Weka default "committee size" or interleave of 3 in our experiments to test our second empirical question: Can Multiboost with Bookmaker Informedness (MultiBook) or Kappa (MultiKap) weighted accuracy achieve better boosting and overcome the hypothesized disadvantage of MultiBook due to the weak learner being optimized for Accuracy?

## 3  Data Sets and Algorithms

For comparability with the other work on multiclass boosting (e.g. Zhu et al. 2009), we use the same character set datasets as shown in Table 1 along with their number of classes, attributes and instances. 2x5-fold Cross Validation was used for all experiments. As we in general had 26 English letters, 26, 260 and 2600 boosting iterations were tested (a weak learner may boost just one class). Graphs for the Multiboost vs Adaboost comparisons show Standard Deviations (red extension bars) and 2 standard error Confidence Intervals (black whiskers).

Table 1. Datasets and Statistics. Tra indicates Training set only used although Test set exists (2x5-CV always used).

| Dataset with K Classes | Attributes | Instances |
|---|---|---|
| Handwritten 10 | 256 | 1593 |
| Isolet 26 | 617 | 7797 |
| Letter 26 | 16 | 20000 |
| OptDigitsTra 10 | 64 | 3823 |
| PenDigitsTra 10 | 17 | 7494 |
| Vowel 11 | 13 | 990 |

We have explored the chance-corrected boosting of Naïve Bayes with results as summarized in Fig. 1. It is noted that only for one data set, Vowel, was significant boosting achieved, and for one, Letter, all the chance-corrected versions made things marginally worse (but not to a degree that is either practically or statistically significant). Also as expected, neither of the chance correction measures was particularly effective, and there was no clear advantage of Booking over Kapping or vice-versa, except that on the one dataset where any boosting happened, Booking was faster than Kapping (with a difference that was only marginally significant at p<0.05 for 26 iterations, and disappeared completely by 260 iterations), but they did not do significantly differently from standard AdaBoost with Rand Accuracy, which actually appeared to be best for this dataset, as well as for Letter as previously noted. Bayesian approaches were thus not pursued further.

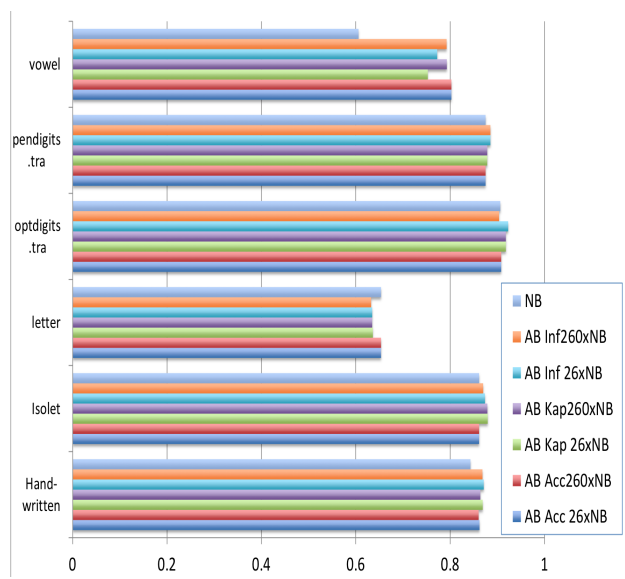

Figure 1. Boosting Naïve Bayes rarely works and chance-correction makes little difference (2x5-CV x 26 or 260 iterations) and we show better results in Figure 3.

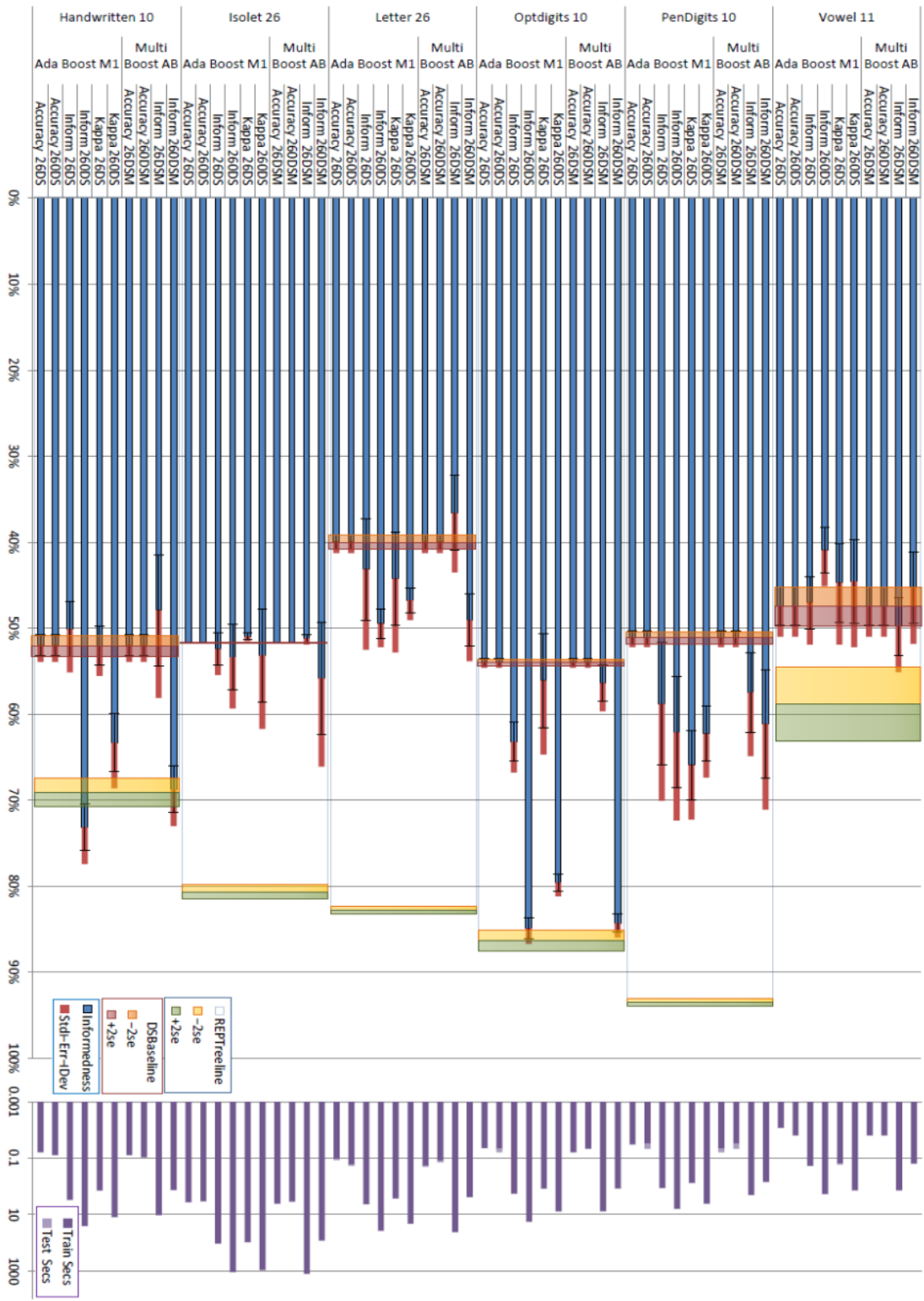

Figure 2. 2x5CVx 26 & 260 iteration AdaBoost & MultiBoost with & without chance correction on Decision Stumps. A 1 standard deviation baseline range (Decision Stump) and treeline range (REPTree) are also shown for reference.

## 4 Results

For weak and strong rule/tree learners where boosting was expected and confirmed (Weka's REPTree and Simple Cart), we tested MultiBoost as well as AdaBoost using Accuracy, Cohen Kappa and Powers Informedness, and selected results are shown in Figs 2 & 3. For reference both the single level and complete Decision Tree are shown as baseline (REPTree Decision Stump) and bestline (REPTree Full Tree). Table 2 also shows 260 cycles for a weak base learner and 26 for 2 strong base learners for additional datasets with SimpleCART as the additional strong base learner.

Although we have tested the family of Perceptron and SVM learners extensively, they are beyond the scope and space of this paper, where for fairness we concentrate on tree-type weak learners as originally proposed for AdaBoost, but we note that similar results pertain. We cannot confirm that MultiBook is better than AdaBook but rather they seem evenly balanced as to which is best: we see no evidence of avoiding early surrender, but suggest that the bagging iterations lead to lower performance on weak learners as less boosting iterations are performed. Matching boosting iterations is a matter for future work, but we see slightly better performance on stronger base learners.

For Decision Stumps (DS in Figs & D260 Table 2) there are two-character recognition cases where MultiBook was significantly better (Vowel and Isolet) and for the other datasets AdaBook seems to be a bit better. In all cases, the uncorrected accuracy versions failed to boost, but boosting was achieved with corrected accuracies. In two of the six cases (Opt and Hand), AdaBook was already comparable with or better than the full REPTree learner, and MultiBook and AdaKap performed slight less spectacularly. It is telling that standard AdaBoost is uncompetitive, and that even with chance-corrected boosting, it mostly fails to attain the REPTree Bestline. In Fig. 2 for both experiments we use 26 and 260 iterations of DS boosting, but in Fig. 3 we show 2600 iterations of AdaBook gives no further gain.

When boosting a stronger REPTree learner (noting that the Decision Stump learner is REPTree restricted to a single branch decision), the story is quite different: in all case all boosting approaches achieved significant improvement over REPTree, with MultiBook apparent best in four of the six cases (similar results for all boosters were achieved for Pen and Opt, but as we approach 100% accuracy, there is less scope to show their mettle, and these had the underlying learners with the highest inherent accuracy). The results for boosting SimpleCART are very similar, and often slightly better than for REPTree as seen in Table 2.

## 5 Conclusions

We have extended chance-corrected adaptive boosting of standard weak learners to include bagging iterations according to the MultiBoost algorithm. Chance correction is found to make a considerable difference to the performance of both AdaBoost.M1 and MultiBoost (with three iterations of AdaBoost.M1 to one of Bagging). Indeed, for a weak learner it tends to make the difference between boosting nicely, and not boosting at all, whilst for a stronger learner, better results tend to be achieved, and no worse results were achieved, except for two of the additional datasets shown in Table 2 where for Sick neither REPTree nor Simple CART showed improvement either, and for Hypothyroid the boosting with Accuracy failed and both Kappa and Informedness regressed the strong learners minor improvement above baseline.

Compared with other variants of boosting or AdaBoost, no inbuilt learning or regression mechanism is required, and no probability or plausibility or confidence rating or ranking needs to be generated for the weak learner: a standard learner can be used and no extension is required. However, it is usually better to start with a strong learner.

Moreover, it is not necessary to run separate training sessions for each class – learning across all classes simultaneously is possible for base classifiers that support this.

On the other hand, boosting performance for Naïve Bayes was spectacularly absent, with only one dataset achieving boosting, and no chance-correction mechanisms showing any advantage versus accuracy. The Naïve Bayes learner is significantly different from a Tree Learners, and this apparent independence may make it suitable for use in multi-classifier variants of boosting, bagging or stacking, based on ensemble fusion techniques involving variation to the classifier rather than just the selection or weighting of data, and using optimization of weights.

A major deficiency of this work is that we used only base learners that were optimized in terms of uncorrected accuracy or error, and it is noted (Powers, 2011) that such optimization can actually make things worse in chance-correct or cost-penalty terms. There is thus a strong chance that the weak learner will detrain and thus not satisfy the boosting condition, and this is particularly likely for Informedness, but less likely for Kappa which is more closely related to Accuracy, and tends more to move with Accuracy, although its divergence from Informedness is itself a source of reduced performance. This explains why often Kappa will seem to do better than Informedness, which should do better on theoretical grounds *given a chance-correct weak learner*.

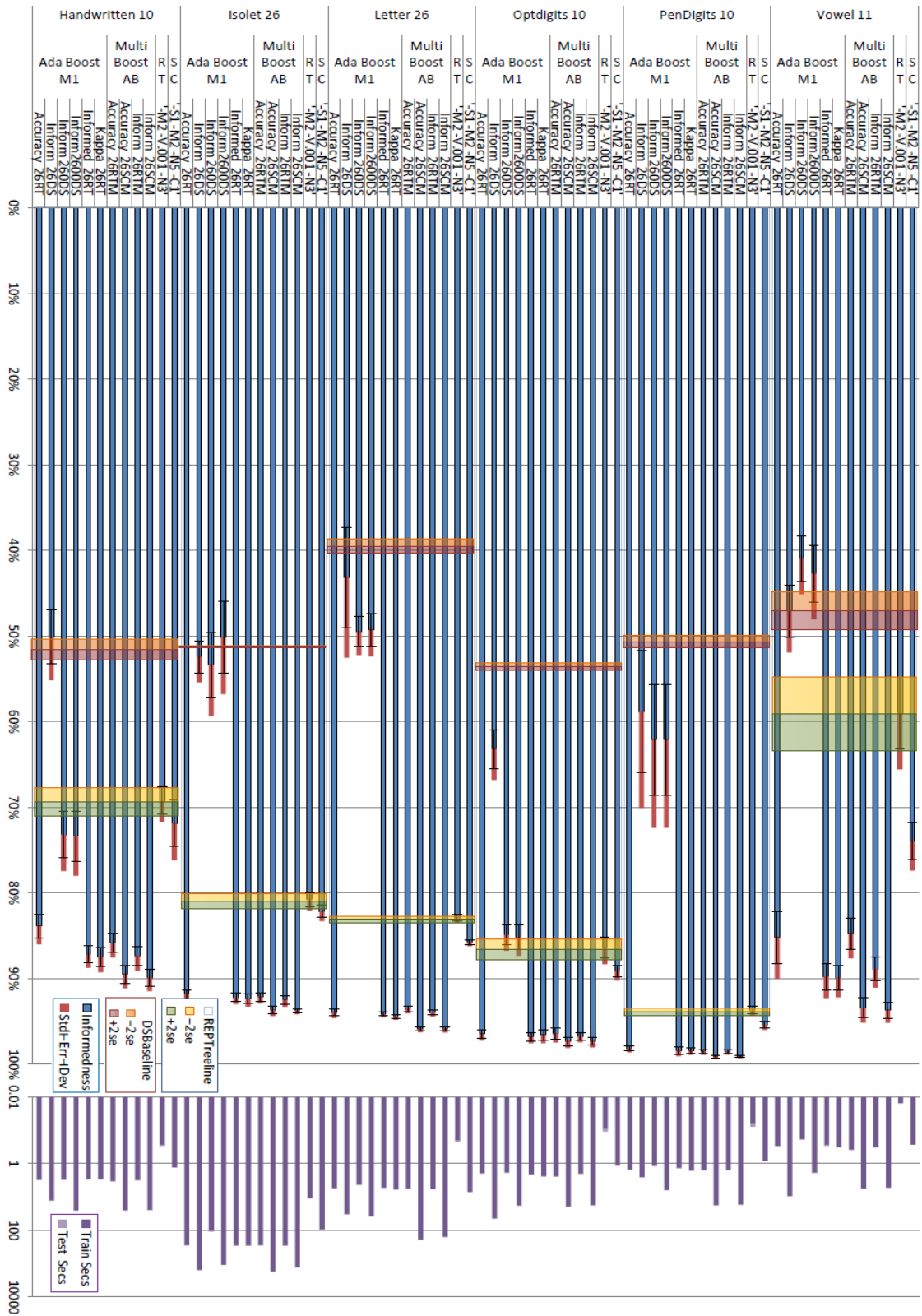

Figure 3. 2x5CVx 26 iteration AdaBoost & MultiBoost with & without chance correction on REPTrees. 2x5CV 26, 260 & 2600 iteration AdaBoost with Informedness, DS baseline, RT bestline shown for comparison (as Fig.2).

Table 2. AdaBoost & MultiBoost by Accuracy, Kappa and Informedness, vs Decision Stump, REPTree & SimpleCART.
Wins within 5% equivalence range are counted, plus Boosts & Losses outside 5% equivalence range round weak learner.
SimpleCART is not shown for space reasons, but is generally slightly better than RT and slightly worse than Boosted SC.
Informedness is shown but if there is a sig. qualitative/directional discrepancy with Accuracy or F1 this is marked AlF.

| Informedness | | | AdaBoost.M1 | | | | Baseline | Treeline |
|---|---|---|---|---|---|---|---|---|
| Dataset | AccRT26B | AccD260B | KapRT26B | KapD260B | InfRT26B | InfD260B | DS | RT |
| Handwritten | **0.84** | 0.52 | *0.87* | 0.56 | *0.87* | **0.73** | 0.52 | *0.69* |
| hypothyroid | *0.99* | 0.38 | 0.97 AF | 0.89 AF | 0.97 AF | 0.84 AF | 0.96 | *0.99* |
| iris | *0.93* | *0.91* | *0.93* | *0.91* | *0.93* | *0.90* | 0.67 | *0.92* |
| Isolet | *0.92* | 0.52 | *0.92* | 0.48 | *0.92* | 0.53 | 0.52 | **0.81** |
| letter | *0.94* | 0.40 | *0.94* | 0.40 A | *0.94* | 0.49 | 0.40 | **0.83** |
| nursery | *0.99* | 0.66 | *0.99* | 0.62 | *0.99* | **0.73** | 0.66 | **0.94** |
| optdigits.tra | *0.96* | 0.54 | *0.97* | 0.69 | *0.97* | **0.85** | 0.54 | **0.86** |
| pendigits.tra | *0.98* | 0.51 | *0.98* | 0.66 | *0.98* | 0.62 | 0.51 | **0.94** |
| segment | *0.96* | 0.53 | *0.97* | **0.80** | *0.97* | **0.81** | 0.53 | **0.94** |
| sick | 0.84 | 0.77 AF | *0.88* | 0.86 | *0.88* | 0.86 AF | 0.86 | 0.84 |
| vowel | **0.85** | 0.47 | *0.90* | 0.41 | *0.90* | 0.41 AF | 0.47 | *0.59* |
| waveform | *0.73* | 0.62 | *0.72* | *0.74* | *0.72* | *0.74* | 0.54 | *0.64* |
| Average | *0.91* | 0.57 | *0.92* | 0.67 | *0.92* | 0.71 | 0.60 | 0.83 |
| EquiWins | 8 | | 9 | 1 | 9 | 1 | 2 | 2 |
| SigBoosts | 10 | 3 | 10 | 8 | 9 | 11 | | 11 |
| SigLosses | | 2 | | | | 1 | | |

| | | | MultiBoost.AB | | | | Baseline | Treeline |
|---|---|---|---|---|---|---|---|---|
| Dataset | AccRT26M | AccSC26M | AccD260M | InfRT26M | InfSC26M | InfD260M | DS | RT |
| Handwritten | **0.86** | *0.89* | 0.52 | **0.87** | *0.90* | *0.69* | 0.52 | *0.69* |
| hypothyroid | 0.98 AF | *0.99* | 0.38 | 0.98 AF | 0.98 AF | 0.84 AF | 0.96 | *0.99* |
| iris | *0.94* | *0.91* | *0.92* | *0.93* | *0.92* | *0.92* | 0.67 | *0.92* |
| Isolet | *0.92* | *0.94* | 0.52 | *0.92* | *0.94* | 0.56 | 0.52 | **0.81** |
| letter | *0.94* | *0.96* | 0.40 | *0.94* | *0.96* | 0.49 | 0.40 | **0.83** |
| nursery | *0.99* | *1.00* | 0.66 | *0.99* | *1.00* | **0.75** | 0.66 | **0.94** |
| optdigits.tra | *0.96* | *0.97* | 0.54 | *0.97* | *0.97* | **0.84** | 0.54 | **0.86** |
| pendigits.tra | *0.98* | *0.99* | 0.51 | *0.99* | *0.99* | 0.61 | 0.51 | **0.94** |
| segment | *0.97* | *0.98* | 0.53 | *0.97* | *0.98* | **0.84** | 0.53 | **0.94** |
| sick | 0.86 | 0.86 | 0.81 AF | *0.87* | *0.87* | 0.86 | 0.86 | 0.84 |
| vowel | **0.85** | *0.93* | 0.47 | **0.89** | *0.94* | 0.45 AF | 0.47 | *0.59* |
| waveform | *0.75* | *0.76* | 0.62 | *0.75* | *0.76* | *0.74* | 0.54 | *0.64* |
| Average | *0.92* | *0.93* | 0.57 | *0.92* | *0.93* | 0.72 | 0.60 | 0.83 |
| EquiWins | 8 | 8 | | 10 | 12 | 1 | 2 | 2 |
| SigBoosts | 8 | 8 | 3 | 9 | 9 | 11 | | 11 |
| SigLosses | | | 2 | | | 1 | | |

Key: Bold Italic represents Maximum; Bold represents better than Treelines (REPTree & SCART);
Italic is near Treelines; Underscore is near Baseline; Strikeout is below Baseline (Decision Stump);
Shading distinguishes underlying learner: REPTree (green) or SCART (lilac) vs Decision Stump (none)

# 6   Future Work

Chance correction has been advocated for decades, but only now is it being incorporated into learners, starting with boosting. It is clear that it should be incorporated into base learners as well (Powers, 2013), and further studies are needed to explore those existing learners that optimize Informedness or some other chance-corrected measure.

In addition, further variations and combinations on boosting, bagging and stacking would seem to be worth exploring to address the limitations of particular weak learners and ensure that boosting is allowed to continue. In particular, techniques that revert to lower K-learners on failure of the weak learner, would gain the best of both worlds – fast multiclass learning where possible, and solid but slow low cardinality or single class learning when not. As noted above, this includes exploring the sensitivity of MultiBoost and its chance-corrected variants to the number of bagging and boosting iterations.

We identify the fact that weak learners are still optimizing an uncorrected measure as the major obstacle to achieving the theoretical performance of chance-corrected boosting, and in Table 2 we have noted with A resp. F cases where the Accuracy res. F1 have risen but chance-corrected measures fell. We are working on general modifications/wrappers for broad classes of learner to address this issue, including a specific focus on ANNs, SVMs and Decision Trees.

It is also a priority to explore boosting of learners that are less sensitive to noise and don't have the convexity constraints of AdaBoost, including learners that are based on switching and can explore and unify alternate learning paths. Since boosting works well with tree learners, such a tree-like approach would produce a consistent but potentially more comprehensible model due to the structural risk minimization properties of boosting and the noise sensitivity minimization properties of switched boosting. However, we are also exploring performance with SVM and MLP (Powers, 2013).

Our focus here was the language/character multiclass problems, but we also have more general problems in robotics, vision, diagnostics etc. However, the diverse natures of these problems, as illustrated by the other half dozen datasets in Table 2, remain to be characterized and understood. It is particularly important to explore what difference chance-correction makes in practical applications, and an obvious application where AdaBoost is a mainstay component, is face finding and object tracking (Viola & Jones, 2001).

This paper has concentrated on two particular kinds of ensemble technique: boosting; and bagging in combination with boosting. One of the explanations of why these techniques work, and why boosting is in general more effective than bagging, is that the different subsets of instances that are selected for learning, and thus the different trained weak learners, correspond to different weightings on the features as well as the examples. Techniques that explore feature evaluation and selection, including ensemble techniques like Random Forests and Feating, more directly select features. When features have different sources (e.g. biomedical sensors, audio sensors and video sensors combined) or have spatiotemporal interrelationships (e.g. pixels or MRI voxels or EEG electrodes sampled at a specific rate), then there is additional structure that may be usefully explored.

AdaBoost, AdaBook and AdaKap may all be used reasonably effectively as early fusion techniques because of these implicit feature selection properties, and in our current work the chance-correction advantage is again clear, although this is beyond the scope of this paper. Nonetheless there seems to be a lot more room for improvement including selecting features and combining weak classifiers in ways that bias towards independence (or decorrelation) rather than using simple majority or convexity fusion techniques as implemented in traditional boosting. This is something else we are exploring.

We have also glossed over the existence of a great many other boosting algorithms, and the known limitations of convex learners such as AdaBoost in dealing with noise. These convex learners are Perceptron-like and the boosted learner is a simple linear combination of the trained weak learners, and are known not to be able to handle label noise. The original boosting algorithms were based on Boolean or voting ideas, and further work is needed on variants of boosting that don't overtrain to noise like Adaboost can, but are insensitive to the occasional bias introduced by label noise, or the regular variance introduced by attribute and measurement noise, or the kind of artefacts and punctuated noise we get in signal processing, including EEG processing, audio speech recognition, and video image or face tracking. The idea is that successive stages bump an instance up or down in likelihood but a mislabelled instance is not repeatedly trained on with increasing weights until labelled "correctly" (Long & Servidio, 2005, 2008, 2010).

We advocate the use of chance-corrected evaluation whenever no actual cost matrix is available. All learning algorithms need to use appropriate costing. Uncorrected/uncosted measures should never be used to across datasets with different prevalences or for algorithms with different biases.

## ACKNOWLEDGEMENTS

This work was supported in part by the Chinese Natural Science Foundation under Grant No. 61070117, and the Beijing Natural Science Foundation under Grant No. 4122004, the Australian Research Council under ARC Thinking Systems Grant No. TS0689874, as well as the Importation and Development of High-Caliber Talents Project of Beijing Municipal Institutions.